\documentclass[letterpaper, 10 pt, journal, twoside]{IEEEtran}

\usepackage[font=small, labelfont=bf]{caption}
\usepackage{multicol}
\usepackage[colorlinks = true, bookmarks=true, pagebackref]{hyperref}
\usepackage{lipsum}
\usepackage{xspace}
\usepackage{todonotes}
\usepackage{amsmath}
\usepackage{amssymb}
\usepackage{booktabs} 
\usepackage{wrapfig}
\usepackage[capitalise]{cleveref}
\usepackage[english]{babel}
\usepackage{here}
\usepackage{flushend}
\usepackage{multirow}
\usepackage{caption}
\usepackage{subcaption}
\usepackage{dsfont}
\usepackage{cite}
\usepackage{siunitx}

\IEEEoverridecommandlockouts                              

\title{\LARGE \bf
Pixel to Elevation: Learning to Predict Elevation Maps at Long Range using Images for Autonomous Offroad Navigation
}

\author{
Chanyoung Chung$^{1,2}$,
Georgios Georgakis$^{1}$, 
Patrick Spieler$^{1}$, 
Curtis Padgett$^{1}$, 
Ali Agha$^{2}$, 
Shehryar Khattak$^{1}$
\thanks{Manuscript received: December 14, 2023; Revised:
February 9, 2024; Accepted: April 7, 2024.}
\thanks{This paper was recommended for publication by
Editor Javier Civera upon evaluation of the Associate Editor and Reviewers’ comments.}
\thanks{$^{1}$ NASA Jet Propulsion Laboratory, California Institute of Technology, Pasadena, CA, USA.}
\thanks{$^{2}$ Field AI, Mission Viejo CA, USA.}
\thanks{The research was carried out at the Jet Propulsion Laboratory, California Institute of Technology, under a contract with the National Aeronautics and Space Administration (80NM0018D0004). This work was partially supported by Defense Advanced Research Projects Agency (DARPA).}
\thanks{\copyright 2024. California Institute of Technology. Government sponsorship acknowledged. All rights reserved.}
\thanks{Digital Object Identifier (DOI): see top of this page.}
}

\begin{document}

\maketitle

\begin{abstract}
Understanding terrain topology at long-range is crucial for the success of off-road robotic missions, especially when navigating at high-speeds. LiDAR sensors, which are currently heavily relied upon for geometric mapping, provide sparse measurements when mapping at greater distances. To address this challenge, we present a novel learning-based approach capable of predicting terrain elevation maps at long-range using only onboard egocentric images in real-time. Our proposed method is comprised of three main elements. First, a transformer-based encoder is introduced that learns cross-view associations between the egocentric views and prior bird-eye-view elevation map predictions. Second, an orientation-aware positional encoding is proposed to incorporate the $3D$ vehicle pose information over complex unstructured terrain with multi-view visual image features.
Lastly, a history-augmented learnable map embedding is proposed to achieve better temporal consistency between elevation map predictions to facilitate the downstream navigational tasks. 
We experimentally validate the applicability of our proposed approach for autonomous offroad robotic navigation in complex and unstructured terrain using real-world offroad driving data. Furthermore, the method is qualitatively and quantitatively compared against the current state-of-the-art methods. Extensive field experiments demonstrate that our method surpasses baseline models in accurately predicting terrain elevation while effectively capturing the overall terrain topology at long-ranges. Finally, ablation studies are conducted to highlight and understand the effect of key components of the proposed approach and validate their suitability to improve off-road robotic navigation capabilities.
\end{abstract}


\section{Introduction}
\label{sec:intro}


Autonomous offroad robotic navigation over unstructured and complex terrains is of considerable interest across a diverse set of robotic application domains, ranging from planetary exploration to agriculture, and search and rescue. In these offroad robotic missions, in particular for high-speed ($\SI{>10}{\meter\per\second}$) traversal, on-board autonomy should be able to reason about the geometric terrain at long-ranges ($\SI{\sim 100}{\meter}$) to efficiently plan trajectories that ensure both safe and optimal navigation. 



Existing off-road autonomous systems primarily rely on LiDAR sensors to navigate the environments by either utilizing maps provided by the Simultaneous Localization and Mapping (SLAM) system~\cite{driveArea,slam2023} or explicitly building them in real-time~\cite{gvom}. 
These maps are then processed by employing heuristics to estimate elevation maps within the mapped space and to understand the traversability of the surrounding area.
However, despite the precision of LiDAR sensors in measuring depth, LiDAR-based approaches suffer from issues associated with the sparsity of LiDAR returns at longer distances. Consequently, this limited perceptual range results in sub-optimal mapping for downstream navigation tasks such as path planning and mission execution. Furthermore, the problem is exacerbated at higher speeds as LiDAR sparsity coupled with larger robot motion between LiDAR scans results in fewer depth measurements per unit area, thus reducing the reliability of the constructed map. 



\begin{figure}[t!]
    \centering
    \includegraphics[width=\columnwidth]{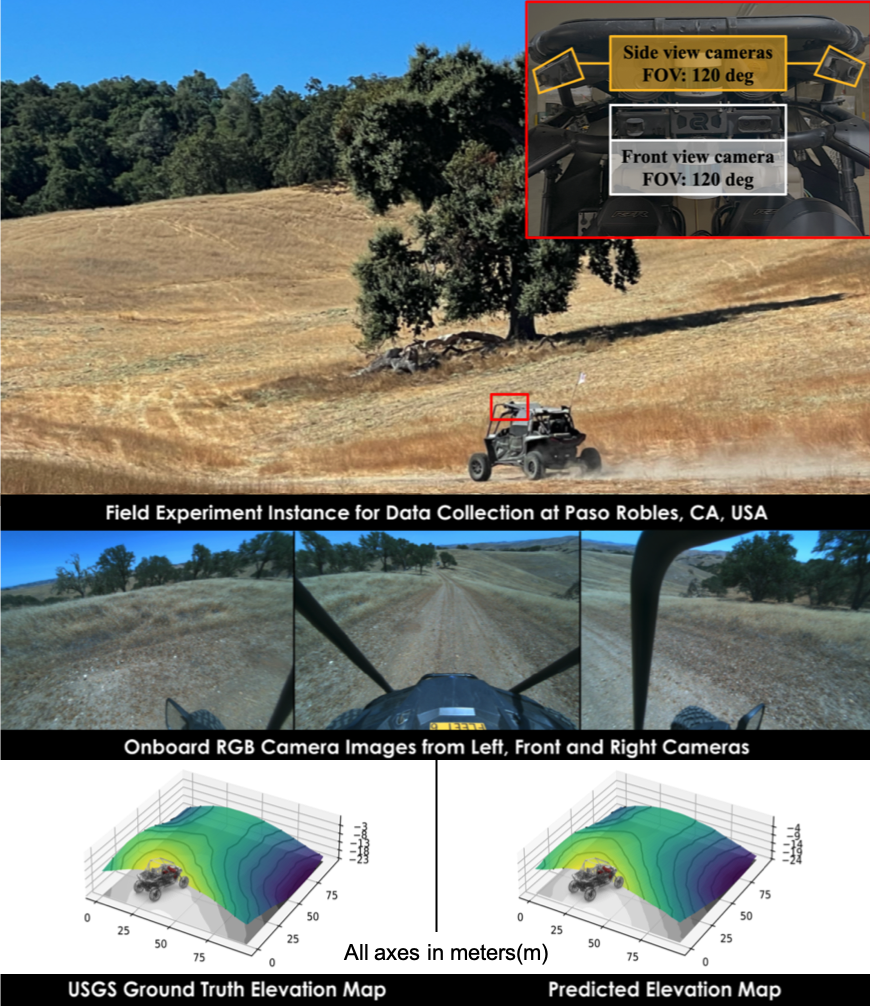}
    \caption{\textbf{Top row} shows the robot navigating in an offroad natural environment during a field experiment conducted in Paso Robles, USA. The red box indicates the mounting position of the visual cameras used in this work, with the inset figure showing a zoomed-in view. \textbf{Middle row} shows the images taken by the left, front and right cameras during an instance of the experiment. \textbf{Bottom row} shows the elevation map output of the proposed method and compares it to the ground truth provided by the USGS. Using only visual camera images as input, our method is able to reliably predict the elevation map up to a distance of $\SI{100}{\meter}$ to facilitate high-speed off-road robot navigation.}
    \label{fig:banner_field}
\vspace{-4ex}
\end{figure}

To overcome these limitations, recent works~\cite{sock2016probabilistic, meng2023terrainnet, krusi2017driving, shaban2022semantic} have proposed to employ learning-based approaches to predict the complete elevation maps at fixed size for offroad and unstructured environments.
Notably, in \cite{meng2023terrainnet}, the author proposed a terrain perception model featuring multiple independent task heads for predicting semantics and 2.5D terrain elevation maps using camera and LiDAR sensor inputs. In \cite{ruetz2023foresttrav}, the Convolutional Neural Network (CNN) base model which takes the 3D voxel representation from the mapping stack and predicts traversability costs within the 3D space is introduced.
However, despite the promising results, the majority of perception models still constrain their map predictions to less than $\SI{50}{\meter}$ in range. In the context of real-world off-road navigation scenarios, extending the perception range further becomes crucial for enabling supporting high-speed driving, proactive planning, and ensuring safe navigation. This imperative drives our motivation to investigate the challenge of generating reliable elevation maps at long-range.

In this letter, we propose a transformer-based Deep Neural Network (DNN) that can predict accurate elevation maps at $\SI{100}{\meter}$ range using only perspective monocular views by leveraging cross-view attention mechanisms. 
The proposed architecture consists of three key components which enable robust elevation map prediction in offroad scenarios:
1) A multi and cross view transformer-based encoder, that can associate multiple perspective views and project visual features into the target $2.5D$ elevation map space,
2) an orientation-aware positional encoding that encodes $3D$ robot pose information over rough unstructured terrains into the learning framework, and
3) a history-augmented map embedding, which improves the temporal consistency of predicted maps for downstream tasks. 

We train and validate the proposed method using real-world off-road data gathered using full-scale off-road vehicles (see Fig. \ref{fig:banner_field}) across diverse environments. Our model is trained in a supervised manner using Digital Elevation Map (DEM) sourced from the publicly available United States Geological Survey (USGS). 
Through extensive quantitative and qualitative evaluations, we demonstrate that our proposed framework improves the elevation map prediction performance over existing state-of-the-art methods. Furthermore, we perform ablation studies to provide insights for our proposed architecture.

The contributions of this letter are as follows:
\begin{itemize}
    \item We propose a learning approach using cross-view attention mechanism to predict accurate and complete elevation maps at long-range ($\SI{100}{\meter}$) in real-time using only onboard perspective camera images as input. 
    \item We propose a novel orientation-aware positional embedding to encode the robot's pose information with the learned visual features, improving the generalization performance when the vehicle travels over rough unstructured terrain. 
    \item We propose a history-augmented map-view embedding, that uses previous prediction recursively to improve temporal consistency across map predictions. 
    \item We conduct a comprehensive experiment evaluation using real-world offroad data, collected using full-scale vehicles across diverse unstructured environments. The results show that our method outperforms the baselines, and emphasize the impact of the key aspects of our proposed method through ablation studies.
\end{itemize}

\section{Related Work}
\label{sec:related_work}

\subsection{Elevation Map from Geometric Mapping}


Early works on elevation map generation \cite{kweon1989terrain, kweon1992high, kolter2009stereo} utilized a feature-matching technique to find the corresponding transformations between multiple sensor measurements, such as stereo cameras or 3D lidar, in order to construct composite elevation maps. Each cell in a 2.5D elevation map stores height information to represent terrain geometry. 
Another widely used approach involves creating elevation maps using voxel maps or 3D point cloud maps, which project 3D data onto a 2D grid structure \cite{fan2021step, hines2021virtual}. 


Most existing elevation map estimation methods heavily rely on depth sensors, including Lidar and stereo cameras. However, these depth sensors often produce sparse information at a distance, resulting in incomplete maps, especially in long-range mapping tasks.





\begin{figure*}[t!]
    \centering
    \includegraphics[width=\textwidth]{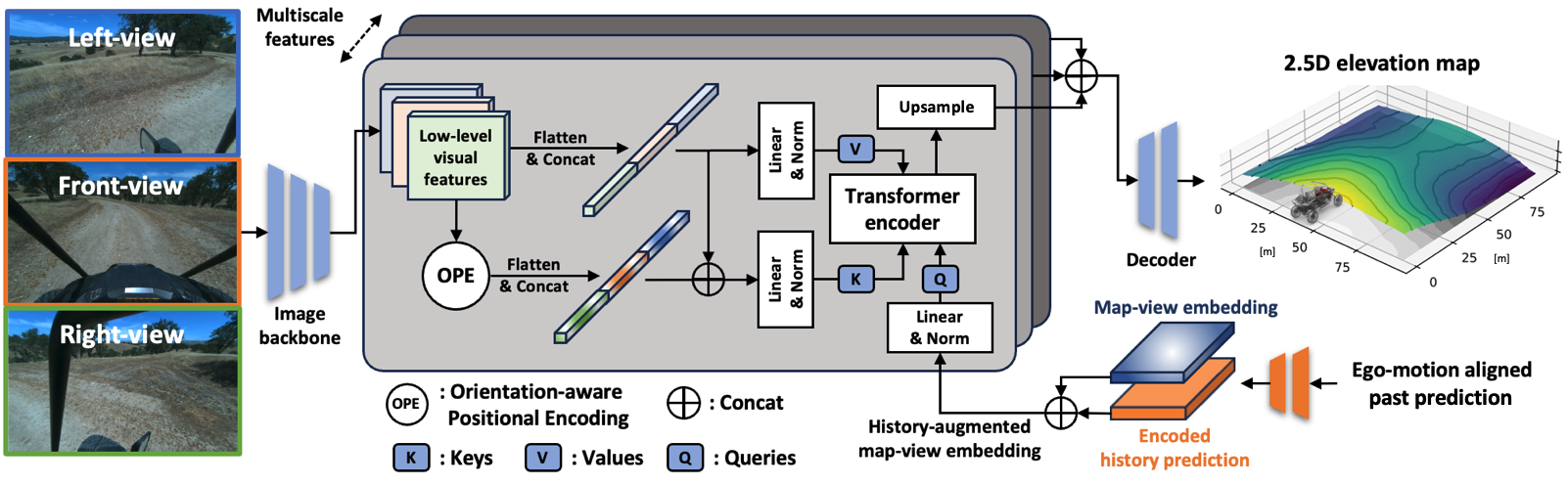}
    \caption{An overview of our proposed architecture for terrain elevation map prediction. Front, left and right camera views are encoded at multiple scales using a shared backbone. The visual features are then transformed into the positional embedding using gravity-aligned vehicle orientation and camera matrix. Our history-augmented map-view query interacts with image features across from multiple views. Output of cross-attention module at each scale is upsampled to the desired dimensions and forwarded to the prediction head.}
    \label{fig:architecture}
\vspace{-2ex}
\end{figure*}

\subsection{Completing Sparse Elevation Map}
In \cite{jaritz2018sparse}, the network that fuse the incomplete RGB-D to accomplish depth completion and semantic segmentation is proposed. Similarly, in \cite{hu2021penet}, densification of the sparse depth image, guided by RGB image using a multi-headed DNN is proposed. Authors aimed to fuse color and depth modalities while respecting scene geometry, resulting in enhanced depth completion accuracy. In \cite{van2019sparse}, authors focused on producing the complete depth map by using sparse and noisy LiDAR and an RGB image. The proposed framework extracts global and local contextual information in a hierarchical manner from the RGB domain and fuses the different modalities into the depth map. 

Learning-based image-impainting are applied to the geometry mapping task. In \cite{qiu2019void}, authors proposed the framework that fills voids in Digital Elevation Models (DEMs). They proposed a deep convolutional generative adversarial network (DCGAN) designed to learn the texture of the surrounding terrain and utilize this understanding to fill in the missing information given DEM. \cite{stolzle2022reconstructing} proposed a neural network that inpaint the occluded 2.5D elevation map with the inputs consisting of the DEM and a binary occlusion mask. Diverging from the aforementioned supervised learning methodologies, they embraced a self-supervised learning mechanism to tackle the case when there is no access to comprehensive and accurate ground-truth data by synthetically generating the missing information using ray-casting algorithms in the map. 

Learning-based 3D reconstruction from images represents a dynamic research domain aimed at inferring missing 3D information. In \cite{hane2017hierarchical}, authors introduced the Hierarchical Surface Prediction (HSP) method, which adeptly reconstructs high-resolution volumetric grids from single-image inputs. \cite{mildenhall2021nerf} presents NEURAL Radiance Field (NeRF), a view synthesis technique leveraging volume rendering with implicit neural scene representation. NeRF has notably achieved state-of-the-art visual quality, showcasing impressive demonstrations and inspiring subsequent works. However, the majority of reconstructions utilizing visual or 3D information are primarily focused on objects or scenes in close proximity.

\subsection{Vision-based Learning Viewpoint Transformations}
Learning viewpoint transformations have been actively proposed which aims to establish connections between the appearances from various perspectives. 
Especially, the map-view, also known as the bird's-eye view (BEV), has been commonly adopted as a canonical space because of its suitability for various downstream tasks 
\cite{ harley2023simple, li2022bevformer, liu2023bevfusion}. 
In \cite{li2022bevformer}, authors proposed semantic segmentation model in BEV space using a transformer-architecture-based DNN which aggregates spatial and temporal information from multiple views. Similarly, in \cite{yang2021projecting}, the authors proposed a DNN that directly predicts road scene layout and vehicle occupancies in BEV space using perspective views. They adopted a cross-attention mechanism \cite{chen2021crossvit} that learns the correlations between different viewpoints to strengthen the view transformation. Also, the context-aware discriminator was proposed to exploit the spatial relationship between the road and vehicles. 

The majority of BEV perception research has concentrated on on-road scenarios, where the flat-plane assumption holds true. However, in off-road situations, this assumption becomes invalid, requiring models to adapt to the diverse postures of vehicles and incorporate this variability into predictions.
\section{Problem statement}
\label{sec:problem_statement}

Our objective is to predict a 2.5D gravity-aligned terrain elevation map, ${}^{t}\mathcal{M}_{i}^{\mathcal{P}_t} \in \mathbb{R}^{H/r \times W/r}$ with respective to the vehicle frame $\mathcal{P}_t$ at timestamp $t$, using the on-board front, left, and right camera monocular views. Here, $i$ denotes the grid cell index. For simplicity, we may omit subscript and superscript unless otherwise specified. 

The dimensions of the prediction area, $H$ and $W$, are both set to 100 meters, with resolution $r$ to 1 meter. We position the vehicle at the bottom-middle of the prediction area and assign terrain elevations to individual grid cells relative to the vehicle's frame of reference. Here, we define terrain elevation as a representation exclusively focusing on the topographical surface of the ground, while disregarding other surface features such as trees and bushes. The prediction of the terrain elevation is formulated as a supervised regression task, where the ground-truth DEM are obtained from the USGS.

\section{Proposed Methodology}
\label{sec:method}

\subsection{Overview}

We propose a method for long-range elevation map prediction that incorporates three important ingredients: 1) learning cross-view associations between multiple image inputs and a map, 2) incorporating vehicle pose information when learning these associations, and 3) including temporal information as encoded in the history of prior predictions.
These key design elements are realized in the form of a multi-scale transformer encoder (Sec.~\ref{subsec:attention}), an orientation-aware positional encoding (Sec.~\ref{subsec:ope}), and a history-augmented map-view embedding (Sec.~\ref{subsec:map_embedding}), respectively.

Given the input images from multiple cameras, we first extract multi-scale visual features using a backbone feature extractor. These features are then concatenated with the orientation-aware positional embeddings and passed to the transformer encoder. The latter also receives the learned map view embeddings concatenated with the encoded history of predictions. The motivation behind this design is to allow the transformer to learn features that associate different views spatially and from multiple time steps into the map-view perspective. Finally, the output features of the transformer are passed into a decoder head that regresses the 2.5D elevation map. The overall architecture of our proposed model is illustrated in Fig. \ref{fig:architecture}.

\subsection{Orientation-aware Positional Encoding}\label{subsec:ope}

Predicting 2.5D elevation maps using egocentric views can be framed as a view transformation task. We introduce the Orientation-aware Positional Encoding (OPE) module to establish proximity between two different views while the vehicle navigates uneven terrain environments by incorporating camera parameters and vehicle pose information.

We begin with unprojecting the $i$-th 2D features $(u_i,v_i)$ into the 3D direction vector as follows:
\begin{equation}
    \vec{d}_i = \mathbf{P} \cdot
    \begin{bmatrix}
        u_i, v_i, 1
    \end{bmatrix}^{\mathsf{T}},
\label{eq:ope}
\end{equation}
where $\mathbf{P} = \mathbf{G}^{-1}\mathbf{R}^{-1}\mathbf{K}^{-1} \in \mathbb{R}^{3 \times 3}$ is a projection function using the camera intrinsic $\mathbf{K} \in \mathbb{R}^{3 \times 3}$, extrinsic $\mathbf{R} \in \mathbb{R}^{3 \times 3}$, and gravity-aligned vehicle orientation $\mathbf{G} \in \mathbb{R}^{3 \times 3}$.

Similar to \cite{li2022bevformer, saha2022translating, zhou2022cross}, we encoded this direction vector into the single vector of positional embedding by passing through the MLP, which we subsequently employed as keys in the following attention mechanism\cite{vaswani2017attention}. However, the difference with the previous works is that we also injected the vehicle's pose information into the visual features as well as the static camera parameters. This simple extension effectively mitigates visual ambiguity caused by the vehicle's pose when navigating uneven terrain.


\begin{figure}[t!]
    \centering
    \includegraphics[width=\columnwidth]{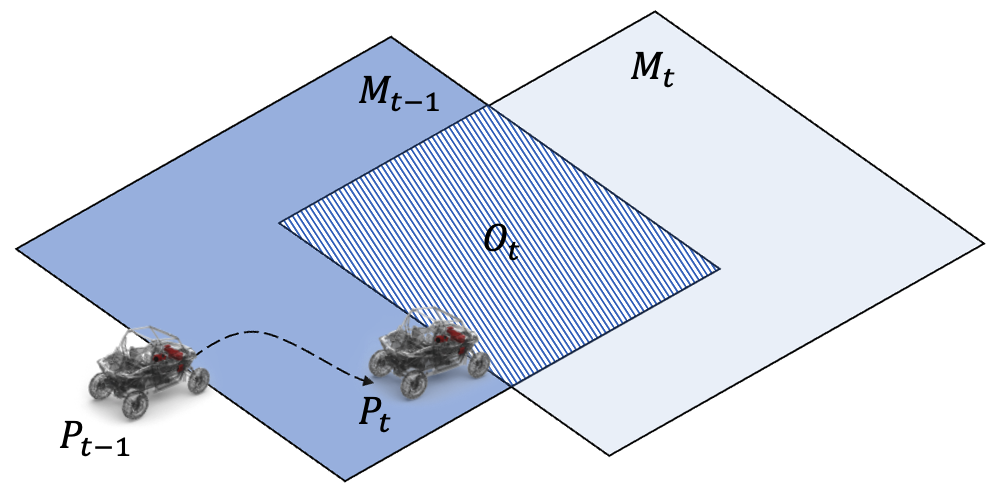}
    \caption{Illustration of prediction area alignment over time for a history-augmented map-view embedding. $\mathcal{M}_t$ represents the elevation map prediction area at timestamp $t$ with respect to robot pose, $\mathcal{P}_t$.  $\mathcal{O}_{t}$ represents the overlap area between predictions $\mathcal{M}_{t-1}$ and $\mathcal{M}_{t}$, represented with respect to pose $\mathcal{P}_t$.}
    \label{fig:ego_motion_aligned}
\vspace{-4ex}
\end{figure}

\subsection{History-augmented Map-view Embedding}
\label{subsec:map_embedding}
We predefined a learnable map-view embedding $Q \in \mathbb{R}^{Q_h \times Q_w \times Q_c}$ similar to \cite{zhou2022cross}, wherein a learnable embedding is initialized as the query to interact with visual features. Each cell is associated with the embedding vector $Q_p \in \mathbb{R}^{1 \times Q_c}$ positioned at $p=(i, j)$, and each embedding vector is used for querying the front-view feature map to generate the corresponding map-view feature. $Q_h$ and $Q_w$ have the same size as the prediction space. 


In addition to the $Q$, we enhanced our queries by explicitly incorporating the previous prediction. This augmentation imbues our queries with both spatial cues and historical information, providing a valuable foundation for achieving temporal consistent prediction. 
To this end, we first aligned the previous prediction, ${}^{t-1}\mathcal{M}^{\mathcal{P}_{t-1}}$, into the current vehicle frame, $\mathcal{P}_{t}$ and we denote this as ${}^{t-1}\mathcal{M}^{\mathcal{P}_{t}}$. We computed the overlapping boolean mask ${}^{t}\mathcal{O}^{\mathcal{P}_t}$, between the ${}^{t-1}\mathcal{M}^{\mathcal{P}_{t}}$ and ${}^{t}\mathcal{M}^{\mathcal{P}_{t}}$. with respective to the $\mathcal{P}_{t}$ (See Fig. \ref{fig:ego_motion_aligned}). Note that we utilized our onboard LiDAR-based localization solution~\cite{rose} for the ego-motion alignment. 

${}^{t}\mathcal{O}^{\mathcal{P}_t} \cdot {}^{t-1}\mathcal{M}^{\mathcal{P}_{t}}$ is then encoded using the simple CNN encoder using two Conv 1x1 layers with feature dim of 16 and 32. The encoded past prediction is then appended to the $Q$ and finally used as the queries. Note that the past prediction and was initialized as zero when it was unavailable. 

\subsection{Cross-View transformer}
\label{subsec:attention}

We employed the transformer encoder with attention mechanisms as the key operation for the elevation map prediction. Specifically, we applied the attention mechanism across multiple views for two reasons. First, it enables the model to consider relationships between pixels or features across various perspectives, leading to a more accurate representation of spatial dependencies. Second, multiple camera views offer a richer spatial context, allowing the model to capture a more comprehensive understanding of the terrain. This is crucial for accurately predicting elevation maps, as the terrain's morphology is linked to the spatial relationships within and between neighboring regions.

To this end, we concatenated all the visual features from multiple views in one single vector and trained the model geometric reasoning between the perspective views and map-view representation. 
The visual features $\phi$ encoded by the shared backbone from multiple cameras undergo flattening and concatenation into a vector. The sequence of concatenation follows the order of front, left, and right views: $\phi \leftarrow [\phi_{f} + \phi_{l} + \phi_{r}]$. Similarly, the vectors of orientation-aware positional embeddings $\delta$ are concatenated in the same order: $\delta \leftarrow [\delta_{f} + \delta_{l} + \delta_{r}]$. 

While it is common to concatenate all input features and then use self-attention, we follow a slightly different route. We treat the concatenated visual features and positional embeddings $[\delta,\phi]$ as the key, the history-augmented map-view embedding as the query, and the visual features as the value and use the typical scaled dot-product cross attention. The idea behind this design is two-fold. First, we directly learn an affinity matrix between the orientation-aware visual features and the map that spatially grounds the observations in the bird-eye view.
Second, the map representation that encodes prior predictions is used to query the current time-step observations resulting in an attended representation over the map regions used for predicting the elevation map.

\subsection{Loss Function}

We adopt the smoothed L1 loss 
between the predicted map $\mathcal{M}$, and ground truth $\overline{\mathcal{M}}$, as our main reconstruction loss, $\mathcal{L}_{recons}$. Inspired from \cite{li2018megadepth}, we add the gradient matching loss $\mathcal{L}_{grad}$ to encourage smoother elevation changes while preserving high frequency features in the output:
\begin{equation}
    \mathcal{L}_{grad} = \frac{1}{|E|}\sum_{i \in E}(\|\nabla_x E_i\|_1 + \|\nabla_y E_i\|_1),
\end{equation}
where $E_i = \overline{\mathcal{M}_i} - \mathcal{M}_i$ at the $i$-th index of the map. We also explicitly penalize the model if it outputs inconsistent predictions across time using temporal consistency loss $\mathcal{L}_{tc}$:
\begin{equation}
    \mathcal{L}_{tc} = \frac{1}{|\mathcal{O}|}\sum_{i \in \mathcal{O}}
    \|{}^{t}\mathcal{O}_{i}^{\mathcal{P}_t} \cdot ({}^{t}\mathcal{M}_{i}^{\mathcal{P}_{t}} -  {}^{t-1}\mathcal{M}_{i}^{\mathcal{P}_{t}})\|_1.
    \label{eq:tc_loss}
\end{equation}
Finally, the total variation regularization \cite{aly2005image} loss, $\mathcal{L}_{tv}$, on the final predicted elevation map was adapted.

The overall training loss function, denoted as $\mathcal{L}_{tot}$, is formulated as a weighted combination of the aforementioned losses:
\begin{equation}
    \mathcal{L}_{tot} = \mathcal{L}_{recons} + \mu\mathcal{L}_{grad} + \lambda\mathcal{L}_{tc} + \gamma\mathcal{L}_{tv}.
\end{equation}

During training, we observed that using a large weight of $\lambda$ makes the training struggle to converge. We keep this weight factor relatively small compared to the other weight parameters and set it empirically.

\subsection{Implementation details}
We rectify and resize all perspective images to $512\times512$. We used the pre-trained EfficientNet-B0 \cite{tan2019efficientnet} as the shared image backbone network and extracted multiscale features at \{4,5,6\} stages.
$\phi$ and $\delta$ are then passed into separate linear projection layers with sizes of $d=\{256, 128, 64\}$ at each feature scale. The map-view queries and output dimensions are configured to be $100\times100$. The feature dimension of map-view queries is specified as $c_d=32$, and the encoded previous prediction is channel-wise concatenated. The outputs of the transformer encoder at each scale are reshaped and upsampled by \{4,8,16\} times. 

We opted for a DEM resolution of $\SI{1}{\meter}$, the highest available, for both training and testing in our study. Additionally, we deliberately selected the most recent DEM from USGS to ensure up-to-date data. To align the DEM with our local coordinate system, we employed our GPS-based localization and lidar-based mapping systems. Through rough transformation using global posture information and matching with our lidar-based mapping, we ensured more accurate supervision signals for our analysis. 
\section{EXPERIMENTS AND RESULTS}
\label{sec:experiment and result}

\subsection{Dataset}
\label{sec:dataset}
We gathered real-world offroad data using a modified Polaris RZR all-terrain vehicle. The vehicle is equipped with four MultiSense S27 stereo cameras (front, front-left, front-right, back), and for this work we utilize RGB images provided by the three front facing cameras. For the OPE, we leverage the vehicle's global position obtained from the onboard localization system~\cite{rose,gmsf}.


\begin{figure}
     \centering
     \includegraphics[width=\columnwidth]{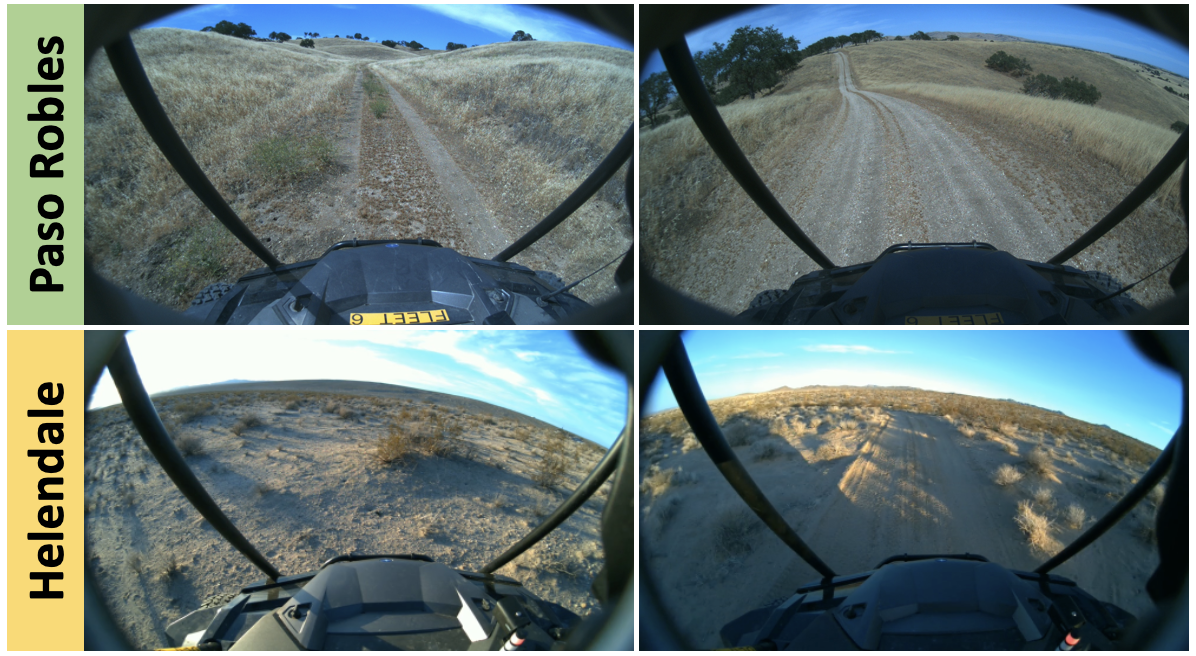}
     \caption{The figure shows examples of onboard camera images taken during field tests in two different environments, highlighting the variety of terrain and test conditions presented. Images from the Paso Robles site (top row) show grassy, hilly terrain with rapid changes in elevation. The Helendale site (bottom row) was a desert terrain containing scattered rocks and bushes.}
     \label{fig:data_sample}
\vspace{-4ex}
\end{figure}

We collected offroad driving data in Helendale and Paso Robles regions in California, USA. As shown in Fig. \ref{fig:data_sample}, Helendale region is part of the Mojave desert and its desert-like landscape is characterized by vast open regions filled with sporadic small bushes with relatively modest shifts in elevation through the environment. In contrast, the Paso Robles region sites exhibit more pronounced changes in terrain elevation, with tall grass and trees covering significant portions of the experiment environment. We used over 6 hours of driving records resulting in around 8,700 training samples, 870 validation samples, and 1,320 test samples.

\begin{figure*}[!t]
    \centering
    \includegraphics[width=\textwidth]{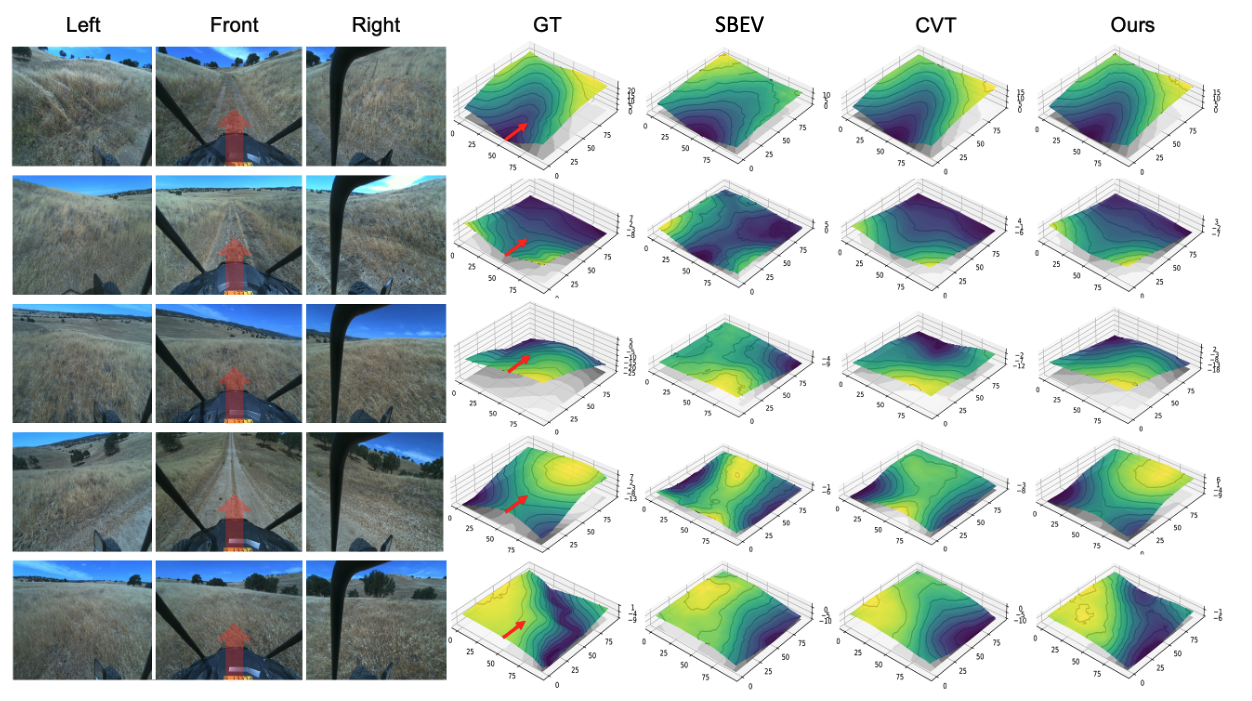}
    \caption{Demonstration of our results on the Paso Robles test set compared with SBEV, CVT, and Ground Truth (GT). The vehicle's direction is denoted by a red arrow, and the elevation at the vehicle's position always corresponds to zero.}
    \label{fig:comparision}
\vspace{-2ex}
\end{figure*}

\subsection{Evaluation metrics and baseline comparisons}
To quantitatively evaluate the proposed method we utilized ground truth data provided by USGS maps. Three quantitative evaluation metrics were employed, namely the Mean Absolute Error (MAE), the Structural Disagreement Rate (SDR), and the mean Temporal Consistency (mTC) \cite{varghese2020unsupervised}. While MAE evaluates the accuracy of predictions for each cell in the map, SDR measures the overall structural consistency between the prediction and ground truth. Furthermore, the mTC indicates the stability of the predictions over time. In line with prior works~\cite{li2018megadepth, chen2016single}, the SDR metric is given as:
\begin{equation}
\begin{aligned}
    SDR(D,\overline{D})&=\frac{1}{n}\sum_{i,j\in D}\mathds{1}(ord(D_i, D_j ) \neq ord(\overline{D}_i, \overline{D}_j)), \\
    ord(&D_i,D_j) =
    \begin{cases}
      1 & \text{if $\frac{D_i}{D_j} > 1 + \tau$,}\\
      -1 & \text{if $\frac{D_i}{D_j} < 1 - \tau$,}\\
      0 & \text{otherwise}.
    \end{cases}
\end{aligned}
\label{eq:sdr}
\end{equation}
where $D$ and $\overline{D}$ refer to the normalized prediction and ground truth elevation maps within the 0 to 1 range, respectively. $(i,j)$ denotes a set of pairs of randomly selected map cell indices, with $n$ representing the sample size. For our evaluation, we randomly sample 100 map cell pairs per prediction and the $\tau$ value is set to 0.1, similar to~\cite{li2018megadepth}.

Furthermore, we compare to relevant works to understand the relative performance of the proposed method. However, given the absence of prior studies directly addressing our elevation map regression task using solely monocular images views, we adapted and compared our approach with representative state-of-the-art architectures designed for urban autonomous driving tasks. MonoL \cite{mani2020monolayout}, SBEV \cite{harley2023simple}, CVT \cite{zhou2022cross}, and LSS \cite{philion2020lift}, were adapted for the elevation map prediction task by replacing their task heads and making minor adjustments to ensure compatibility with prediction sizes. To provide a fair comparison for approaches that use only a single camera, performance evaluation is exclusively performed within the map area enclosed by the camera frustum. Finally, we also trained a single camera version of our proposed method (Ours-F) for evaluation purposes and to understand the performance difference of using a single camera as compared to multiple cameras. Note that, all methods, including ours, were trained and tested using identical setups.

\subsection{Quantitative and qualitative results}

\def\arraystretch{1.1}
\begin{table}[t!]
\caption{Comparison results between multiple baseline models for elevation map prediction. “F”, “L”, and “R” indicate the Front, Left, and Right cameras, respectively. "PR" and "H" indicate the data from Paso Robles and Helendale, respectively, and "PR+H" indicates a combination of PR and H data.}
\centering
\label{tab:comparison}
\begin{tabular}{ccccccc}
\hline
\multirow{2}{*}{Method} & \multirow{2}{*}{Input} & \multirow{2}{*}{Set} & \multicolumn{3}{c}{MAE [m] $\downarrow$} & \multirow{2}{*}{SDR $\downarrow$}  \\ \cline{4-6} &     &       & 0-25m    & 0-50m & 0-100m  \\ \hline 

MonoL               & F     & \multirow{6}{*}{\rotatebox[origin=c]{90}{PR+H Validation}}               & 0.91  & 1.22 & 1.45 & 0.181 \\
LSS                 & FLR   &   & 0.81 & 1.11 & 1.35 & 0.153 \\
SBEV                & FLR   &   & 0.83 & 1.13 & 1.49 & 0.162 \\
CVT                 & FLR   &   & \textbf{0.78} & 0.99 & 1.21 & 0.131 \\
Ours-F              & F     &   & 0.82 & 1.23 & 1.43 & 0.173 \\
Ours                & FLR   &   & 0.8  & \textbf{0.93} & \textbf{1.18} & \textbf{0.128} \\ \hline

MonoL               & F     & \multirow{6}{*}{\rotatebox[origin=c]{90}{H Test}}                        & 0.43  & 0.91 & 1.79 & 0.17  \\
LSS                 & FLR   &   & 0.4  & 0.84 & 1.68 & 0.144 \\
SBEV                & FLR   &   & 0.41 & 0.82 & 1.63 & 0.131 \\
CVT                 & FLR   &   & \textbf{0.38} & \textbf{0.79} & 1.61 & 0.117 \\
Ours-F              & F     &   & 0.42 & 0.88 & 1.93 & 0.166 \\
Ours                & FLR   &   & 0.39 & \textbf{0.79} & \textbf{1.58} & \textbf{0.108} \\ \hline

MonoL               & F     & \multirow{6}{*}{\rotatebox[origin=c]{90}{PR Test}}                        & 1.26  & 2.04 & 3.59 & 0.315 \\
LSS                 & FLR   &   & 1.13 & 1.84 & 3.03 & 0.304 \\
SBEV                & FLR   &   & 1.13 & 1.72 & 2.99 & 0.289 \\
CVT                 & FLR   &   & 0.83 & 1.81 & 2.94 & 0.251 \\
Ours-F              & F     &   & 1.27 & 2.23 & 3.3  & 0.313 \\
Ours                & FLR   &   & \textbf{0.80} & \textbf{1.65} & \textbf{2.74} & \textbf{0.223} \\ \hline
\end{tabular}
\vspace{-4ex}
\end{table}

Quantitative results are listed in Table~\ref{tab:comparison}. MonoL and the proposed Ours-F models, which use a single front-view, exhibit noticeably higher prediction errors compared to models using multiple views across all evaluation sets. In particular, we observed a more significant performance degradation in the Paso Robles (PR) test set compared to the Helendale (H) test set for both methods. This suggests that utilizing a broader observation of the environment could enhance the model's understanding of terrain elevation in general, proving particularly crucial in environment with rapid elevation changes. 

Methods utilizing multi-camera images show accurate predictions on both MAE and SDR metrics on the validation set containing images from both the Paso Robles (PR) and the Helendale (H) regions.
For close (0-25) meter range, CVT showed slightly better performance than ours in terms of MAE, however, the proposed method shows best performance for the longer-range cases i.e. the 0-50 and 0-100 meter ranges. Furthermore, the proposed method also produced the most consistent results for the validation set as shown by the SDR metric in Table~\ref{tab:comparison}.
Evaluation results on the H test set show a similar trend with the proposed method showing better elevation performance at longer ranges and producing more structurally consistent maps, as noted by the lowest SDR error.
Results using the PR test set reveal that every method, including ours, produce larger errors in both metrics compared to the H test set. As mentioned in Section \ref{sec:dataset}, the PR test set features more rapid elevation changes, making it a more challenging environment than the H test set. The proposed method outperformed the compared methods in both MAE and SDR for the full range evaluation case of 0-100 meters, with the most significant performance gap observed on the PR test set in the SDR metric. This suggests that our method effectively captures the contextual information of terrain morphology from visual observations and accurately integrates them into the long-range map view space to produce consistent maps. 


Fig.~\ref{fig:comparision} shows a qualitative comparison of the key experimental results within the PR test set for the 3 best performing methods (SBEV, CVT, and ours). In the first and second rows, the vehicle navigated through a ditch while ascending and descending slowly, respectively. Our model and CVT successfully predicted the left and right slopes while accurately capturing the overall incline. However, SBEV struggled to predict the consistently descending slope in the second-row scenario.
In the fourth row case, the vehicle drove along the flat trail approaching a hill ahead. The proposed method accurately captured the terrain elevation geometry, whereas other methods failed in predicting the large but gradual variation in elevation.
The fifth row shows a particularly challenging scenario where sharp negative ditches lie ahead of the vehicle which remain imperceptible from the camera images. Humans can anticipate such obstacles by comprehending contextual clues or visual features, such as partially visible tree tops across the field and the discontinuity along the ground. Although, our method reasonably predicted the negative slope, it did not perfectly replicate high-frequency features present in the ground truth map. Nevertheless, our model successfully identified the ditch throughout the entire predicted area by discerning visual features of occlusions from multiple viewpoints. Finally, for suitability towards real-time applications, during inference our proposed approach is able to achieve 41 FPS on an Nvidia RTX 3080 Ti GPU.


\def\arraystretch{1.1}
\begin{table}[t!]
\caption{Ablation results.}
\centering
\begin{tabular}{c|cc|cccc}
\hline
Test set                &  OPE & HA & MAE [m] & SDR & mTC [m]\\ \hline
\multirow{4}{*}{H test} &      &                           &1.66     &0.121     &0.53     \\
                        & \checkmark     &                 &1.63     &0.118     &0.48     \\
                        &      &\checkmark                 &1.62     &0.112     &0.42     \\
                        & \checkmark     &\checkmark       &1.58     &0.108     &0.32     \\ \hline
\multirow{4}{*}{PR test} &      &                           &3.07     &0.266     &0.81     \\ 
                        & \checkmark     &                 &2.81     &0.251     &0.77     \\
                        &      &\checkmark                 &2.93     &0.249     &0.60     \\
                        & \checkmark     &\checkmark       &2.74     &0.223     &0.51     \\ \hline
\end{tabular}
\label{tab:ablation_overall}
\vspace{-4ex}
\end{table}

\subsection{Ablation study}
To understand the contribution of each component on the overall performance of the proposed method, an ablation study was conducted. In particular, the inclusion of orientation-aware position encoding (OPE) and history-augmented (HA) map-view embedding modules were evaluated, with quantitative results are presented in Table~\ref{tab:ablation_overall}.

\subsubsection{Effectiveness of OPE}
We compared our OPE module with the camera-aware positional encoding (CPE) method proposed in \cite{zhou2022cross}. Briefly, CPE projects visual features using only camera intrinsic/extrinsic parameters without considering the vehicle's pose. In Table \ref{tab:ablation_overall}, we applied CPE instead when the OPE was not applied. Regardless of the test set, OPE improved the performance as demonstrated by lower MAE and SDR values. Notably, the performance improvement was relatively larger in the PR test set which contains more challenging terrain. 

\begin{figure}[t!]
    \centering
    \includegraphics[width=\columnwidth]{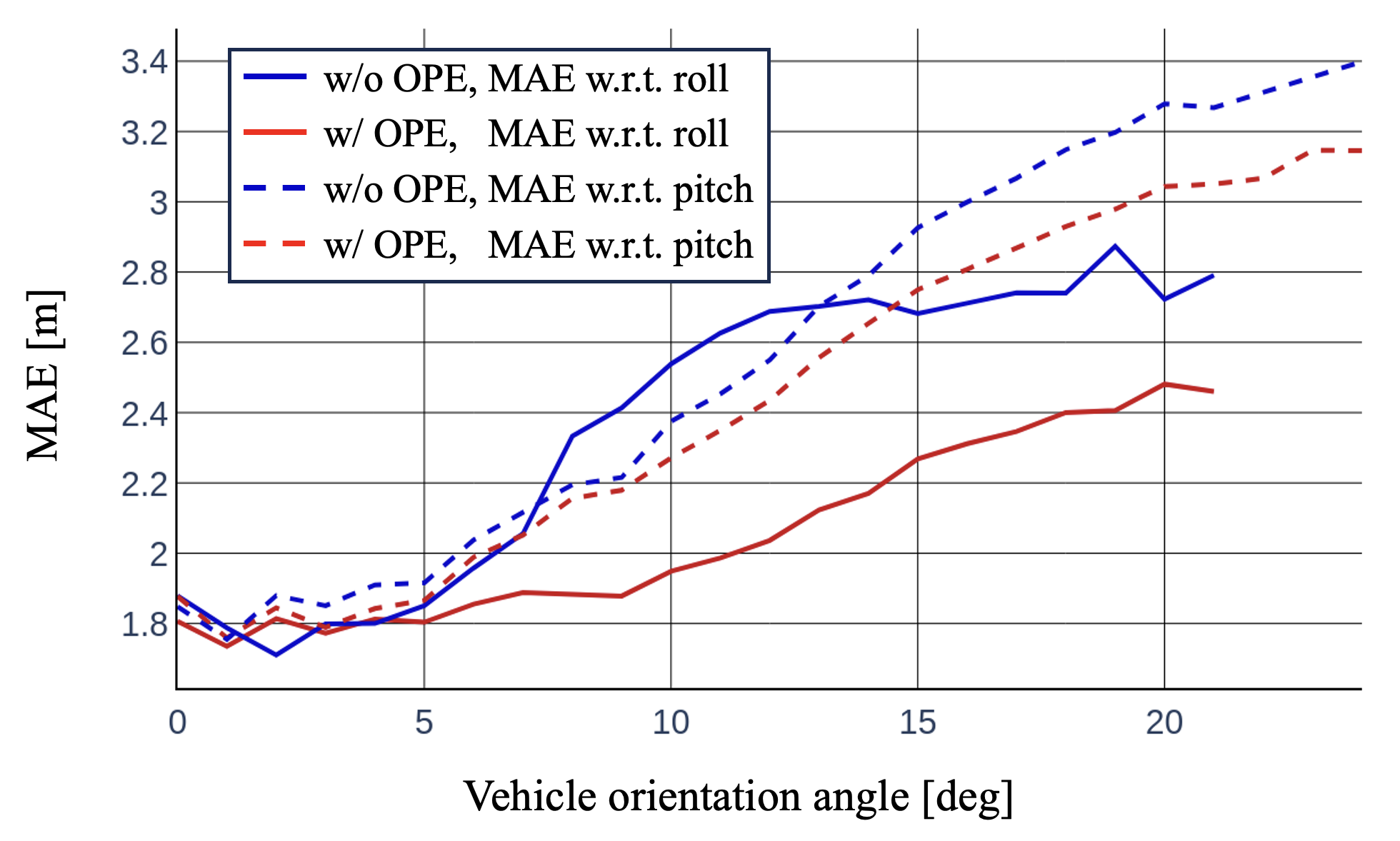}
    \caption{The plot shows the effect of vehicle orientation on elevation map prediction error and the benefit of adding Orientation-aware Positional Encoding (OPE). The solid and dashed lines show the effect of roll and pitch,respectively.}
    \label{fig:ablation_ope}
\vspace{-4ex}
\end{figure}

To further demonstrate the utility of OPE, we examined test sets featuring high roll and pitch angles. Illustrated in Fig. \ref{fig:ablation_ope}, it can be noted that prediction errors increase in all cases with the vehicle orientation angle increase, in both roll and pitch. For pitch angles, OPE slightly improves performance over CPE. However, OPE outperforms CPE for large roll angles. We hypothesize that explicitly integrating gravity-aligned roll and pitch rotations via OPE can alleviate the ambiguity of visual features, offering valuable geometric cues to the model for improved accuracy in predicting the terrain elevation. 



\subsubsection{Effectiveness of HA}
We compared our history-augmented map-view embedding to the learnable map-view embedding proposed in~\cite{zhou2022cross}. Aligned with the objective of improving temporal consistency between predictions using history-augmented map-view embedding both prediction accuracy and the temporal consistency error (Eq~\ref{eq:tc_loss}) were evaluated. 

Table \ref{tab:ablation_overall} shows that using history-augmented map-view embedding module marginally improved performance both in terms of MAE and SDR, irrespective of the presence of the OPE module. In particular, for the H test set, HA improved the prediction accuracy in terms of MAE about $\SI{0.05}{\meter}$ and $\SI{0.04}{\meter}$ with and without OPE, respectively. For the PR test set, slightly higher improvement of $\SI{0.07}{\meter}$ and $\SI{0.14}{\meter}$ was seen with and without OPE, respectively. Notably, the majority of the performance improvements are seen in temporal consistency metric. In H test set, we see improvements of $\SI{0.16}{\meter}$ and $\SI{0.11}{\meter}$ with and without OPE, respectively. Similarly, for the PR test set, improvements of $\SI{0.21}{\meter}$ and $\SI{0.26}{\meter}$ in the TC metric with and without OPE, respectively, were reported. 

The best overall performance was achieved when both OPE and HA were employed concurrently, constituting our proposed approach. Through ablation studies, we contend that OPE serves as a simple yet effective positional encoding method, particularly valuable when the vehicle navigates uneven terrain, contributing to accurate elevation map predictions. Additionally, our proposed HA module, while making a modest contribution to accuracy, plays a substantial role in ensuring consistent predictions -- a crucial requirement for downstream tasks such as planning.



\section{Conclusion}
\label{sec:discussion}

In this letter, we proposed a novel learning-based long-range elevation map prediction approach using onboard monocular images only for offroad navigation. As part of our design, the orientation-aware positional encoding module incorporates the vehicle orientation into visual features which improves the prediction accuracy especially when the vehicle navigates over rough and uneven terrains. Additionally, our history-augmented map-view embedding recursively encodes previous predictions with the spatial map-view queries, providing temporal cues to the model for consistent map predictions. 
Extensive field experiments and ablation studies show that our model outperforms existing baselines and contributions of each component. 


While the proposed approach has shown promising results using only monocular views for long-range elevation map prediction, there are still areas for improvement that we aim to address in our future research to enhance both performance and practicality. One such avenue for improvement is the incorporation of depth images or Lidar as additional input modalities, which could help reduce range uncertainty and improve precision. Additionally, achieving consistent performance across different environments remains a challenging task. In this regard, we believe that estimating the uncertainty of our predictions could be beneficial for downstream tasks such as planning and control.





\bibliography{references}
\bibliographystyle{IEEEtran}

\end{document}